\documentclass[a4paper,12pt]{article}
\usepackage{lmodern}
\usepackage{abstract}
\usepackage{pdfpages} 
\usepackage{booktabs}
\usepackage{placeins}
\usepackage[flushleft]{threeparttable} 
\usepackage[margin=1in]{geometry}
\usepackage{amsfonts} 
\usepackage[colorinlistoftodos]{todonotes}
\usepackage{longtable}
\usepackage{tabularx}
\usepackage{array}
\usepackage{rotating}
\usepackage{verbatim}
\usepackage{setspace}
\usepackage{amssymb}
\usepackage{amsmath}
\usepackage{xcolor}

\usepackage[T1]{fontenc}
\usepackage{floatrow} 
\usepackage{textcomp} 
\usepackage{natbib}
\usepackage{graphicx}
\usepackage[permil]{overpic}
\usepackage{multirow}
\usepackage{amsmath, amsthm, amssymb} 
  \numberwithin{equation}{section} 

\usepackage{accents} 
\usepackage{savefnmark}
\usepackage{paralist}
\usepackage{caption} 
\usepackage{subcaption}
\usepackage[affil-it]{authblk}
\usepackage{blindtext}
\usepackage[english]{babel}
\usepackage[utf8]{inputenc}
\usepackage{tcolorbox}
\usepackage{import}

\usepackage{chngcntr}
\counterwithout{equation}{section}

\usepackage[raiselinks, pdfborder={0 0 0}, bookmarksopen=true, bookmarksnumbered]{hyperref}

\hypersetup{hidelinks,
urlcolor=blue
}

\begin{document}

\title{
    Empirical Privacy Evaluations of Generative and Predictive Machine Learning Models\\
        \Large{A review and challenges for practice}\thanks{This work was supported by Open Data Infrastructure for Social Science and Economic Innovations.}\\
}

\author{\textsc{Flavio Hafner}\thanks{%
Netherlands eScience Center, \texttt{f.hafner@esciencecenter.nl}}
\textsc{Chang Sun}\thanks{%
Institute of Data Science, Maastricht University, \texttt{chang.sun@maastrichtuniversity.nl}}
}

\date{ \today }

\maketitle
\vspace{-10mm}
\begin{abstract}

   Synthetic data generators, when trained using privacy-preserving techniques like differential privacy, promise to produce synthetic data with formal privacy guarantees, facilitating the sharing of sensitive data. However, it is crucial to empirically assess the privacy risks associated with the generated synthetic data before deploying generative technologies. This paper outlines the key concepts and assumptions underlying empirical privacy evaluation in machine learning-based generative and predictive models. Then, this paper explores the practical challenges for privacy evaluations of generative models for use cases with millions of training records, such as data from statistical agencies and healthcare providers. Our findings indicate that methods designed to verify the correct operation of the training algorithm are effective for large datasets, but they often assume an adversary that is unrealistic in many scenarios. Based on the findings, we highlight a crucial trade-off between the computational feasibility of the evaluation and the level of realism of the assumed threat model. Finally, we conclude with ideas and suggestions for future research.

\end{abstract}

\section{Introduction}
Research in the social and health sciences often relies on accessing tabular data sourced from statistical agencies or healthcare providers. These datasets, containing sensitive personal information, are typically subjected to stringent regulations to protect privacy. Synthetic personal data - designed to comply with privacy protection regulations and standards - can address these challenges by enhancing data accessibility and supporting the reproducibility of scientific work and statistical studies~\citep{mukherjee2023UtilityForRCTs}. 

The development of deep generative models has introduced significant capabilities in generating realistic synthetic data, which mimic real datasets while protecting sensitive information. This approach provides a promising solution to enhance data accessibility and support the reproducibility of scientific work and statistical studies where data privacy is crucial~\citep{mukherjee2023UtilityForRCTs}. One approach to creating privacy-preserving synthetic data is using deep generative models trained with the Differentially Private Stochastic Gradient Descent algorithm (DP-SGD, \citet{abadi2016deep}), which gives theoretical privacy guarantees~\citep{dwork2013AlgorithmicFoundations}. However, the effectiveness of these techniques in protecting the privacy of original data has not yet been thoroughly investigated. It is crucial to empirically assess the privacy risk of a trained privacy-preserving generative model for several reasons.


First, before synthetic data generators can be deployed in practice, they must gain the trust of data owners such as statistics offices and healthcare providers \citep{cummings2023CenteringPolicy}. For these stakeholders, the intuitive complexity of the DP-SGD algorithm often exceeds that of traditional statistical disclosure constraints or simpler DP mechanisms, such as those used to release aggregate data. Without accessible explanations and demonstrations of its effectiveness, these stakeholders may hesitate to adopt such methods.

Second, DP-SGD does not account for adversaries with access to auxiliary datasets, which can increase privacy risks. This gap necessitates additional evaluations to assess vulnerabilities in real-world scenarios
 \citep{cummings2023Challenges}. For example, an adversary might use external datasets to reconstruct sensitive information from synthetic data, which is a scenario not fully addressed by DP-SGD’s theoretical framework.

Third, while the theoretical bounds of DP-SGD are derived under the assumption of a highly capable adversary, they are often conservative in practical applications. This conservatism suggests that a generative model might achieve higher statistical utility if the privacy guarantee is relaxed to align with more realistic adversarial scenarios. Empirical evaluations can bridge this gap by assessing privacy leakage under practical conditions, enabling stakeholders to better balance privacy and utility.


Privacy attacks are an essential tool for empirically quantifying privacy leakage, as they simulate attempts to infer sensitive information from data or model outputs~\citep{10.1145/3624010}. Privacy attacks attempt to infer sensitive information from the data or model outputs, allowing us to assess the extent to which privacy may be compromised. Several software packages were developed to conduct privacy attacks on generative models \citep{houssiau2022Tapas, qian2023synthcity} or predictive models  \citep{kumar2020mlprivacy}, establishing a set of evaluation metrics to assess the privacy-preserving capabilities of generative models.

However, one primary challenge for interpreting the outcomes of these evaluations requires a robust understanding of the underlying methodologies and assumptions. This is especially critical as empirical evaluations approximate theoretical privacy protections, often requiring significant computational resources to produce actionable insights. For practitioners and newcomers to privacy research, these complexities present a considerable barrier to entry. Moreover, it is an open question to what extent (large-scale) privacy testing should and could be relied upon in practice \citep{jagielski2020Auditing, yoon2020anonymization}.

Second, the evaluation of these privacy-preserving models (including generative and predictive models) lacks a universally accepted or standardized framework, leading to inconsistencies in methodologies and criteria. For instance, studies on generative adversarial networks (GANs) vary widely in their approach to privacy evaluation. Some do not include any privacy assessment \citep{fang2022dp, xie2018differentially}, while others evaluate privacy by measuring the distance to the nearest synthetic record \citep{xu2018synthesizing, sun2022CorrelationCapture, choi2017generating} and a few utilize sophisticated techniques such as membership inference attacks  \citep{park2018synthesis}. While some of these studies predate the development of the methods we discuss, the lack of standardization makes it difficult to compare the privacy guarantees of candidate algorithms before they are deployed in practice.




This paper focuses on the mechanics of empirical privacy testing, complementing existing reviews~\citep{cummings2023Challenges, ponomareva2023DPfy} with a detailed exploration of differential privacy’s practical implications for generative models. This work is related to the discussion of privacy testing in \citet{houssiau2022Tapas}, combined with the recent literature on direct empirical tests of differential privacy - called privacy audits \citet{ponomareva2023DPfy} - and highlighting practical challenges for using the tests in practice. We look into privacy testing both for predictive and generative machine learning models. While methods for testing predictive models have advanced more rapidly than those for generative models, we believe the former holds valuable lessons for the latter. Additionally, we discuss novel suggestions for advancing privacy evaluations for generative models to address gaps in the literature that predominantly focus on predictive machine learning models. This paper targets a broader audience ranging from statistical officers and researchers in the health and social sciences to researchers and practitioners who are actively engaged in privacy research. We also review recent advancements in the field, highlighting computationally feasible methods while addressing implementation flaws in privacy-preserving algorithms for both generative and predictive models \citep{nasr2023TightAuditing, stadler2022Groundhog}. Finally, we propose several ideas for future work to address the challenges and enhance the practicality of such attacks.

\section*{Concepts and Definitions}
In this section, we introduce the key concepts and definitions in the paper including differential privacy, Stochastic Gradient Descent with Differential Privacy, and the hypothesis testing interpretation of differential privacy. 

\paragraph{Differential privacy (DP)}\label{def:differential-privacy}~\citet{dwork2013AlgorithmicFoundations} is defined as A randomized algorithm $\mathcal{M}$ is $(\varepsilon, \delta)$-differentially private if for all $\mathcal{S} \subset Range(\mathcal{M})$ and for all neighboring databases $D$ and $D^{\prime}$ that differ by at most one record:
    \begin{equation*}
        \Pr [\mathcal{M}(D) \in \mathcal{S}] \leq \exp (\varepsilon) \Pr [\mathcal{M}(D^{\prime}) \in \mathcal{S}] + \delta
    \end{equation*}
where the probability space is over the coin flips of the mechanism $\mathcal{M}$.

Intuitively, this means that the distributions over all the outcomes $\mathcal{S}$ are similar when the input data sets $D$ and $D^{\prime}$ only vary slightly---by one record. Higher privacy is associated with a lower $\varepsilon$: It provides an upper bound on the ability to distinguish the output of $\mathcal{M}(D)$ from the output of $\mathcal{M}(D^{\prime})$; this bound fails to hold with probability $\delta$.
Further, a key property of differential privacy is that it is immune to post-processing \citep{dwork2013AlgorithmicFoundations}: any operation performed on the output of $\mathcal{M}$ has no worse privacy guarantees than the output of $\mathcal{M}$.


Some extensions to standard DP allow one-to-one correspondences to the definition provided here. For instance, both Renyi-DP \citep{ponomareva2023DPfy} and cases of Gaussian DP \citep{dong2019Gaussian, nasr2023TightAuditing} are convertible to $(\varepsilon, \delta)$-DP. There is a well-known trade-off between privacy and statistical utility: the higher the level of privacy protection, the more noise is added to the original data and therefore the less useful they are to draw statistical inferences from. 


\paragraph{Differentially private predictive and generative models}
Both predictive and generative machine learning models could leak sensitive information about the training data: either from the model's predictions, or from the synthetic data produced by the generative models. 
For neural networks, research efforts have focused on developing approaches to train these models with differential privacy aiming to safeguard the privacy of the training data \citep{abadi2016deep}. 
Currently, the main approach for training neural networks with DP is \emph{Differentially Private Stochastic Gradient Descent (DP-SGD)}. It works by clipping the gradients of each individual sample to a maximum norm, and infusing Gaussian noise to the aggregated batch-level gradient. This yields the exact Gaussian mechanism at the batch level. However, for computing the privacy loss from an end-to-end training algorithm, it is necessary to aggregate the privacy loss from each training step, also accounting for the randomization of records into different batches in each epoch \citep{nasr2023TightAuditing}.


DP-SGD works with assumptions on the adversary that are often not discussed explicitly, but it has implications for how the results of privacy audits are interpreted: The adversary sees all gradients and parameter update steps during training; these are used to calculate the algorithms' level of privacy protection.

\paragraph{The hypothesis testing interpretation of differential privacy}

To connect the theoretical guarantees of DP-SGD to empirical measures of privacy, one can interpret DP from a statistical hypothesis testing perspective \citep{wasserman2009StatisticalFramework,kairouz2015CompositionTheorem}. 
Specifically, given an output $Y$ from a randomized mechanism $\mathcal{M}$, consider the two hypotheses:

\begin{align} \label{eq:hypothesis-testing}
\begin{split}
  H0 &: Y \text{ was drawn from } \mathcal{M}(D) \\
  H1 &: Y \text{ was drawn from } \mathcal{M}(D^{\prime}) 
\end{split}
\end{align}

If $\mathcal{M}$ satisfies $(\varepsilon, \delta)$-DP, then an attacker attempting to distinguish between the two hypotheses faces a trade-off between type-I error rate $\alpha$ and type-II error rate $\beta$:

\begin{align*}
 \alpha + e^{\varepsilon} \beta &\geq 1 - \delta \\ 
 e^{\varepsilon} \alpha + \beta &\geq 1 - \delta
\end{align*}


Thus, if one can empirically conduct this hypothesis test, and given a value for $\delta$, it is possible to estimate a lower bound on $\varepsilon$, which is sometimes called \emph{effective epsilon}. 
There are two methods to obtain the effective epsilon. First, the system of inequalities implies
\begin{equation*}
    e^\varepsilon(\delta) \geq \max \left( \frac{1 - \alpha - \delta}{\beta}, \frac{1 - \beta - \delta}{\alpha} \right)
\end{equation*}
Second, the system also implies an upper bound on the accuracy of a binary classification problem where 
\begin{equation*}
    \text{acc}(\delta) \leq \frac{e^{\varepsilon } + \delta}{1 + e^{\varepsilon}}
\end{equation*}

Since the lower bound is derived for a specific attack, future attacks with higher accuracy could be developed, potentially revealing greater privacy leakage of $\mathcal{M}$. 
Moreover, as with any statistical inference, estimating a lower bound for  $\varepsilon$ also needs to account for sampling uncertainty. This is typically done with Clopper-Pearson confidence intervals \citep{crow1956confidence}.

The hypothesis testing interpretation is the foundation for membership inference attacks and privacy audits, which we describe in the following sessions.



\section*{Privacy attacks in models and applications}

This section first introduces the principles of privacy attacks and empirical privacy testing. Although research on auditing generative algorithms is under development, there are significant conceptual overlaps in privacy attacks in predictive and generative models. Therefore, we will differentiate between them only when necessary. Thereafter, we discuss practical challenges to using these methods in practice.

Privacy attacks on machine learning applications empirically apply the hypothesis testing interpretation of differential privacy. \citet{shokri2017MembershipInference} were the first to run a membership inference attack against a machine learning model. The membership inference attack involves determining whether a given data sample was part of the training dataset of a model ~\citep{irolla2019demystifying}. 
Such attacks assess whether an adversary, under assumptions with varying degrees of realism, can infer information about the training data. Based on this foundation, membership inference attacks targeting generative machine learning have been developed. Comprehensive reviews of these attacks are provided by \citet{hu2022MIASurvey} and \citet{chen2020GAN-Leaks}.

\subsection*{Building blocks of attacks}

Privacy attacks simulate an adversary that tries to break the promise of privacy of algorithm $\mathcal{M}$. Although we leave the output of $\mathcal{M}$ unspecified, typically for predictive models it is a vector of predicted probabilities or a scalar of the predicted class. For generative models, it is typically a new dataset.
Simulating the adversary requires certain assumptions that may not be realistic in all contexts. 

\paragraph{An attack target}
The notion of "neighboring" databases requires that there is one record that differs between two otherwise identical databases. The differing record is the attack target. 

\paragraph{Knowledge about the training data.}
The adversary needs data to learn to distinguish whether outputs of $\mathcal{M}$ were drawn from one input database or from another. Thus, the adversary has at least a random sample of records that are drawn from the same distribution as the training data of the original model $\mathcal{M}$.

\paragraph{Knowledge about the algorithm $\mathcal{M}$.}
In all cases, the adversary has at least query access to $\mathcal{M}$. In a predictive model, this means they can make predictions with $\mathcal{M}$ on new datan a generative model, this means they can use $\mathcal{M}$ to create a new synthetic data set of any size. This is the first counter-intuitive assumption for attacks on synthetic data where, in a realistic scenario, not the trained model but only a synthetic data set is released. To our knowledge, such a threat model is only considered in \citet{xu2022MACE}.

The second important counter-intuitive assumption is that the adversary knows the architecture of $\mathcal{M}$, or at least can train a new model with the same architecture. The assumption is explicitly discussed in early work \citep{shokri2017MembershipInference} but not in more recent work. This assumption is stronger in some contexts than in others. For instance, in \citet{shokri2017MembershipInference} the adversary attacks the prediction of an online machine learning service, and can therefore train a new model with the same architecture by using the service. 
In the context of releasing synthetic personal data, however, it is less obvious that the adversary knows how the generator was trained, yielding a stronger adversary than is perhaps realistic. The strongest adversaries know all the gradients from training $\mathcal{M}$ and the resulting parameters. This is referred to as "white-box model access", and corresponds to the threat model assumed to derive the privacy guarantee of DP-SGD.

\subsection*{Shadow modeling and attack logic} 
The adversary wants to find out whether the target record $x_0$  was included for training the released algorithm $\mathcal{M}$. To do so, the adversary starts with a random sample from the population data distribution without the target record---the baseline training set. Then, they draw a random bit $b_t \in \{0, 1\}$. If $b_t=1$, the adversary adds the target record to the baseline training set.
The resulting dataset is then used to train $\mathcal{M}_t$.
Doing so $T$ times results in a set of pairs $\{(b_t, \mathcal{M}_t)\}_{t=1}^{T}$. 

The adversary now uses the set of pairs to learn the decision rule $\mathcal{B}$. Intuitively, the more the inclusion of the target record impacts the gradients during training of $\mathcal{M}$, the more different the learned model $\mathcal{M}_t$  with $b_t = 0$ from models with $b_t = 1$. It thus becomes easier to distinguish the models with $b_t = 0$ from models with $b_t = 1$.
There are different ways to obtain $\mathcal{B}$, and they differ between generative and predictive models.

To learn $\mathcal{B}$  for predictive models, the idea is that $\mathcal{M}$ performs differently on samples that were used for training than on samples that were not used for training, and this is measured through the record's loss. 
For a review of attacks against predictive and generative models, we refer the reader to \citet{hu2022MIASurvey}. We instead focus on insights from \citet{carlini2022FirstPrinciples} and \citet{ye2022Enhanced}. Both studies trained a single threshold to attack multiple training records in a predictive model. Earlier attacks derived one decision rule $\mathcal{B}$ for multiple target records. Such attacks inform about the average privacy risk of a record in the training data but are not informative about the privacy risk of an individual data record. For the latter, one has to train an attack against that particular record, in other words, the target record $x_0$ needs to be fixed across shadow models $(\mathcal{M}_1, ..., \mathcal{M}_T)$.

Even when holding the target record fixed, the adversary can face different sources of uncertainty  \citep{ye2022Enhanced}: There is not only randomness from the starting parameters of $\mathcal{M}_t$ (through the model's seed), but also from the training data besides the target record. In other words, each shadow model may or may not be trained on the same training data (up to the target record). 
The more uncertainty, the weaker the adversary and therefore the higher privacy appears as measured through the effective $\varepsilon$.

Furthermore, the emerging state-of-the-art for membership inference attacks is the likelihood ratio (LR) test \citep{carlini2022FirstPrinciples}. LR tests provide the highest probability of correctly rejecting the null hypothesis in Equation (\ref{eq:hypothesis-testing}) at fixed error rates, and thus the strongest possible adversary given the threat model. However, LR tests require estimating the distribution of losses of models both trained with and without $x_0$ in the training data which are more computationally demanding than computing type I and type II error rates on a hold-out set. Some research work has optimized the computation to solve this challenge such as \citet{zarifzadeh2023LowCostHighPower}.


\subsection*{Membership inference attacks against tabular data synthesizers}

Membership inference attacks against generative models for tabular data use shadow modeling to create a set of synthetic shadow datasets, labeled by whether the target record was included in the training data.
The decision rule $\mathcal{B}$ is then trained to distinguish between two types of synthetic shadow datasets.
The simplest attack compares the distance between the target record and the closest record in the synthetic shadow datasets. Given a distance metric, the attacks learn a threshold, below which the target record is predicted to be included in the training data.
\citet{stadler2022Groundhog} propose a stronger attack by projecting the synthetic dataset onto a lower-dimensional feature space. The simplest way to project is by summarizing statistics and histograms of marginal distributions. $\mathcal{B}$ is then trained on these lower-dimensional features. The python library TAPAS \citep{houssiau2022Tapas} implements both of these attacks and allows to measure the individual privacy risks.

\subsection*{Privacy audits}

Privacy audits \citep{jagielski2020Auditing} are a special case of the hypothesis testing interpretation of DP. 
Privacy audits are designed to maximize the attacker's ability to distinguish between the two hypotheses with the lowest possible type I and type II errors. Thus, privacy audits design the strongest adversary as being free to choose not only the neighboring datasets but also the target record. The adversary can inject any kind of target record into $\mathcal{M}$ and it does not have to be drawn from the original data. This is in line with the definition of differential privacy since it needs to hold for \emph{any} two neighboring datasets, and the differing record is not specified. Some privacy audits refer to this record as a \emph{canary}.

Privacy audits have two motivations. First, membership inference attacks tend to find values for effective epsilon that are lower than the theoretical guarantee derived from DP-SGD. This could be either because the assumed adversary is too weak, or because the accounting of privacy loss in DP-SGD is too conservative, and DP-SGD effectively protects privacy better than what can be proven theoretically. 
If this was the case, one could train DP-SGD with a higher theoretical epsilon, without increasing effective epsilon beyond the acceptable level, and achieve higher statistical utility. In contrast, finding an adversary for whose attack the effective epsilon coincides with the theoretical upper bound would prove that the privacy guarantee derived for DP-SGD is tight. 

The second motivation for privacy audits is to check whether DP-SGD is implemented correctly by finding an effective $epsilon$ above the theoretical lower bound that would identify an error in DP-SGD.

\cite{nasr2021AdversaryInstantiation} 
demonstrated that the privacy guarantees provided by DP-SGD are tight. Using a worst-case adversary, their audit was able to recover the theoretical upper bound on the privacy loss parameter, $\varepsilon$. However, this result relied on an adversary working with a highly unrealistic pathological dataset, namely an empty dataset. In subsequent work, \cite{nasr2023TightAuditing} improved their work and showed that even an adversary with a realistic data set can extract only slightly less information than what is theoretically the upper bound. Because DP-SGD is composed of applying \emph{Gaussian} differential privacy multiple times, they exploit the results from \cite{dong2019Gaussian} to audit the learning mechanism with Gaussian DP. This allows them to audit the algorithm in a computationally more efficient way than previous work.



\subsection*{Practical challenges and suggestions for future research}

Based on previous literature and experience, this section summarizes our insights and opinions on the privacy evaluation methods for generative models and proposes potential future research ideas.

\paragraph{Membership inference attacks with realistic adversaries do not scale to worst-case analysis}

Differential privacy should be upheld for all potential target records in the training data. Thus, a credible test should report privacy guarantees for the record at the highest risk of being identified by an attack, which may require a check for privacy leakage for each record in the training data. 
Performing such a check has a time complexity of 
\begin{equation*}
    N \times \left( T \times O_{\mathcal{M}}(N) + O_{\mathcal{B}}(T) \right)
\end{equation*}
where $N$ is the size of the raw data, $T$ is the number of shadow models trained per target record, and  $O_{\mathcal{M}}(N)$ and $O_{\mathcal{B}}(T)$ denote the time complexity of the generator and the attack algorithms, respectively. 
Even with linear-time training algorithms, this implies a minimum quadratic complexity in the size of the raw data. Consequently, such an assessment is often computationally impractical---not only for real-time deployment \citep{cummings2023Challenges}, but also for comparing generative algorithms on datasets with millions of rows, such as records from registry data. 
While \citet{steinke2023OneRun} propose a method to attack multiple records in a single training run, their attack does not use the likelihood-ratio test. 


One may consider running the attack only on selected target records rather than the whole dataset. For instance,  \citet{stadler2022Groundhog}, \citet{houssiau2022Tapas} and \citet{nezhad2023PrivacyAssessment} selected the target records based on some measure of being an outlier. However, identifying outliers based on the loss function in a model trained on the complete dataset \citep{houssiau2022Tapas} may not reliably identify records with elevated privacy leakage: What matters is the difference in loss between a model trained with the target record included and one without it. In other words, a record may be inherently difficult to fit, leading to a high loss regardless of whether the target record was part of the training data \citep{carlini2022FirstPrinciples}. 

Alternatively, one could make a conscious choice about which records should be protected under a more realistic threat model, train the model accordingly with a lower theoretical privacy guarantee, and run the privacy attacks only on those selected records. However, this approach risks leaking privacy for other records whose vulnerability is not tested. Non-parametric models such as generative adversarial networks \citep{stadler2022Groundhog} learn an implicit density function, making it difficult to know in advance which data dimension the model will represent. This means that the vulnerability of a record that is an outlier on a particular dimension may go unnoticed by this privacy attack. 

Lastly, even if a method were available to efficiently identify the training records with the highest risk of privacy leakage after a model has been trained, it would need to do so in a mathematically rigorous and provable manner


\paragraph{Privacy audits of DP-SGD could be adopted at the level of the algorithm}

In contrast to membership inference attacks, we believe privacy audits involving strong adversaries are useful for ensuring that a particular algorithm correctly implements DP-SGD. The audit method detailed in \citet{nasr2023TightAuditing} requires two training runs, offering significantly improved time complexity compared to membership inference attacks with an adversary with black-box knowledge of the generative algorithm.

Privacy audits may be necessary in practice because there are several reasons for why DP-SGD may not be correctly implemented \citep{nasr2023TightAuditing}: 
First, at the level of each parameter update step, several issues can arise: (i) gradients might not be clipped at the individual sample level, (ii) the added noise may not be generated randomly, and (iii) the noise scale may not be proportional to the batch size, particularly when gradients are computed across multiple machines. 

Second, auditing the entire training across all batches and epochs involves approximating the trade-off function tracing out the adversary's lowest achievable type II error for all type I error rates \citep{dong2019Gaussian}---generated by DP-SGD with the method from \citet{koskela2019TightGuarantees}. However, because records are subsampled into minibatches during training, and the adversary does not see which records are in which sample, this approximation can overstate the actual privacy (leading to a lower bound on $\varepsilon$ that is not tight) of the algorithm, especially at higher false positive rates. This means it may not detect all implementation bugs from the end-to-end perspective, suggesting that such audits should adopt a threshold with a low false positive rate.

These audits operate at the level of a training algorithm, and their results do not depend on the training data  \citep{nasr2023TightAuditing}. Consequently, they are analogous to an integration test in software engineering, which is essential for ensuring the software functions as intended. This is why we believe that automatic and easy-to-use privacy audits on training algorithms are an important tool for using differentially private synthetic data in research praxis.

\paragraph{Making black-box membership inference attacks against generative models feasible}

We believe that membership inference attacks with black-box model access complement those with white-box access as well as privacy audits---precisely because the former depends on context, they are easier to communicate to data owners and give more realistic estimates of privacy leakage. Further, they promise to deliver higher statistical utility than relying on unnecessarily high privacy guarantees from white-box threat models. 
To overcome the aforementioned challenges, we suggest three directions for future research and exploration.

\paragraph{Tailor Threat Models to Specific Contexts}
Each membership inference attack should be customized to its specific use case, with the assumptions tailored to the potential adversary. As a result, establishing a universal benchmark for comparing different generative models is impractical. Instead, the choice of a generative algorithm should be based on the particular use case. If no existing attacks are suitable for a given scenario, new ones must be developed to ensure context-specific effectiveness. 

\paragraph{Employ the strongest adversary for each threat model}
For any given threat model, it is crucial to use the most capable adversary possible. This includes adapting the likelihood ratio test, commonly used for predictive models, to evaluate attacks against generative models. This approach is feasible when the generator has a unit-level loss function as seen in generative adversarial networks. \citet{hayes2017logan} is an early example of using per-example loss of the discriminator in a GAN as data for an attack. This approach could be combined with the attack in \citet{carlini2022FirstPrinciples} to develop a membership inferance attack against generative models that is possibly stronger than existing attacks.

\paragraph{Solve Scalability Issues in Non-Parametric Models}
To empirically test the privacy leakage of generators with black-box threat models, the scalability problem of current attacks needs to be solved. This problem could be tackled by solving two of its underlying sources. 
First, generators with more predictable behavior than non-parametric models may be easier to test for privacy leakage, as it would be clearer which data dimensions are learned by the generative algorithm.  Work in this direction includes \citet{houssiau2022FrameworkAuditableGeneration} which explores creating more auditable generative models.
Second, new training approaches and architectures that incur the large fixed cost of training only once could be promising. For instance, training a a foundation model using public data, and then fine-tuning it with private data for a specific use case could be effective. This approach has been considered for language models  \citep{tramer2022Considerations}, but for population and health data it is unclear if existing publicly available data---such as census tables at geographic disaggregations---contain enough information to do this.



\section*{Conclusion}

We synthesize the literature on empirical privacy testing and discuss some practical challenges with using these tests in practice. 
Our first conclusion is that there is a need to develop workflows and software that make privacy audits easy to use in the learning pipeline. This could foster the adoption of such audits in the machine learning community and is an important first step towards a wider use of DP-SGD in applications.

Our second conclusion is that membership inference attacks with black-box threat models are useful because they provide context to a particular release of synthetic data. But before they can be used more widely in applications, it may be necessary to overcome the scaling problem associated with them, which we consider an important area for future research. 


Our third conclusion is a need for dialogue and research about whether and how privacy audits and privacy attacks should be used in practice.
Empirically quantifying privacy leakage is an important aspect to consider before deploying privacy-protective machine learning systems, possibly alongside other aspects \citep{cummings2023CenteringPolicy, cummings2023Challenges}.
However, relying only on the results from privacy audits makes such systems less useful in practice because the strong adversarial assumption implies low statistical accuracy. 
An important question for future research is therefore whether and which sacrifices on privacy leakage are acceptable to make DP-SGD suitable for opening data and models trained on sensitive data.

\bibliography{main}

\end{document}